\documentclass[letterpaper, 10 pt, conference]{ieeeconf}  

\IEEEoverridecommandlockouts                              
\usepackage{graphics} 
\usepackage{epsfig} 
\usepackage{times} 
\usepackage{amsmath} 
\usepackage{amssymb}  
\usepackage{bm}
\usepackage{bbm}
\usepackage{color, soul}
\usepackage{graphicx}
\usepackage{caption}
\usepackage{subcaption}
\usepackage{cite}
\usepackage[table,xcdraw]{xcolor}

\DeclareMathOperator*{\argMin}{arg\,min}
\title{\LARGE \bf
Modelling and Learning Dynamics for Robotic Food-Cutting}

\author{Ioanna Mitsioni$^{1}$, Yiannis Karayiannidis$^{2}$ and Danica Kragic$^{1}$ 
\thanks{$^{1}$ Division of Robotics, Perception and Learning (RPL), CAS, EECS, KTH Royal Institute of Technology, Stockholm, Sweden 
        {\tt\small mitsioni,dani@kth.se}}%
\thanks{$^{2}$Division of  Systems and Control, Dept. of Electrical Engineering, Chalmers University of Technology, Gothenburg, Sweden 
        {\tt\small yiannis@chalmers.se}}%
}

\begin{document}
\maketitle
\thispagestyle{empty}

\begin{abstract}
Data-driven approaches for modelling contact-rich tasks address many of the difficulties that analytical models  bear. For real-world scenarios, the hardware capabilities constrain the available measurements and consequently, every step of the problem's formulation. In this work, we propose a formulation that encapsulates knowledge from a baseline controller for the contact-rich task of food-cutting. Based on this formulation, we employ deep networks to model the dynamics within a model predictive controller. We design a training process based on curriculum training with learning rate decay for multi-step predictions, which are essential for receding horizon control. Experimental results demonstrate that even with a simple architecture, our model achieves consistently good predictive performance on known and unknown object classes and exhibits a good understanding of the long term dynamics.

\end{abstract}

\section{Introduction}
Modelling and learning dynamics for contact-rich manipulation is an open problem in robotics. Classical control approaches \cite{hogan, Siciliano:2000:RFC:555628, schutter, tune, adaptiveRev, doors, parallel, hybrid} suffer when the modes of interaction increase, their respective models are too complicated to be described analytically or their variations too diverse to be accounted for. Especially for contact-rich tasks, this difficulty arises from the dynamics that can include discontinuities such as breaking and making contact, complicated frictional phenomena, or the variety of object properties. With the introduction of data-driven methods, a lot of these shortcomings were confronted successfully \cite{Kroemer2019ARO}. Their main advantage stems from not relying on analytical models, but on interaction with the environment, or demonstrations that can initialize a policy for the completion of the task.

\begin{figure}[t]
    \centering
    \includegraphics[width = \linewidth]{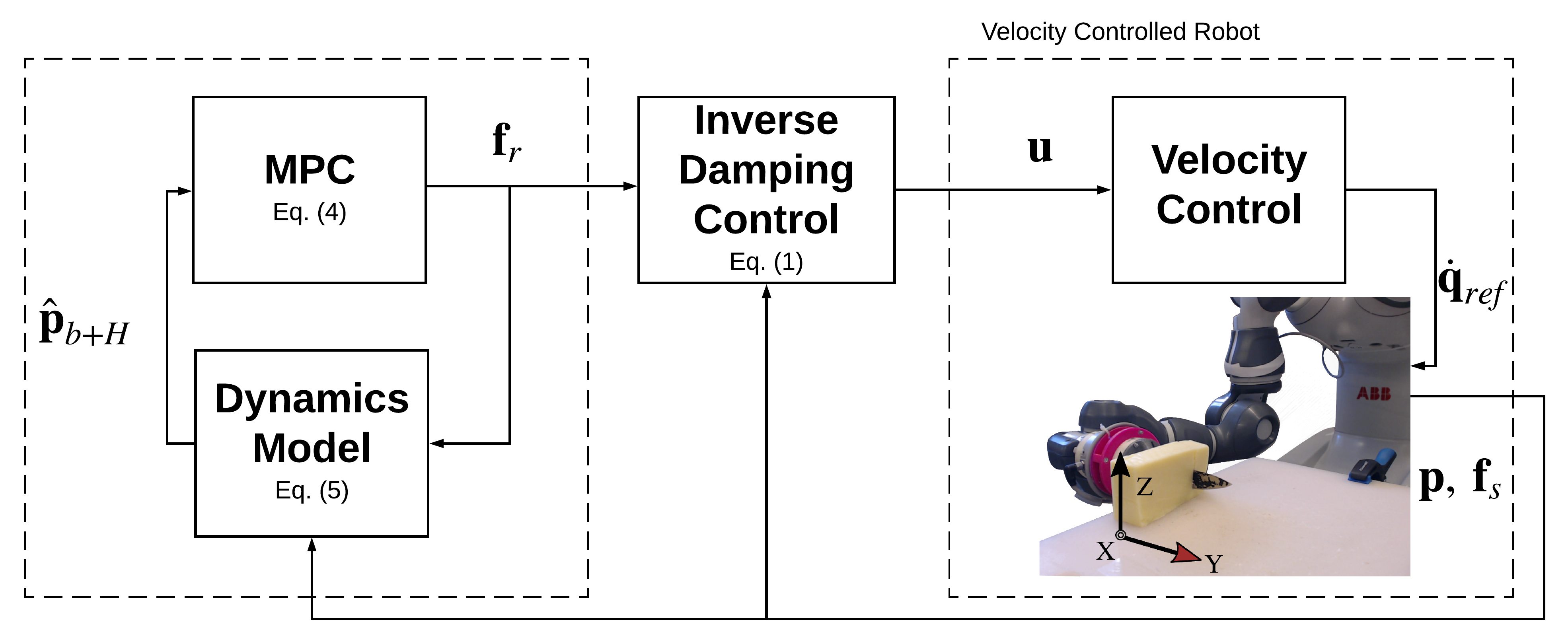}
    \caption{\footnotesize System overview of the method. Axis $Y$ corresponds to the sawing motion required to break friction and enable the downwards motion along axis $Z$.}
    \label{system}
\vspace{-0.3cm}
\end{figure}

In this work, we investigate a data-driven method for robotic food cutting which is inherently a contact-rich task with complicated interaction dynamics. Modelling the interaction as a mass-spring-damper system is an oversimplification of the contact dynamics and the tissue fracturing/separation of the fibers is not well-approximated by a smooth impedance in closed-form expression. On the other hand, more realistic and analytical representations \cite{ATKINS2005479, Mu2019, Long2014} are arduous to develop when considering many different classes, or classes with substantial variations in their dynamics. As a task, cutting can be low-dimensional if only operational space quantities are considered. The discontinuities it exhibits are in most cases due to frictional, stick-slip phenomena and not extreme ones such as sudden and complete breaking of contact. However, it is exceptionally difficult to simulate, so any type of data collection or exploration must be done on the real system, which is expensive. As a result, we choose to learn only the dynamics model from data with a deep network and handle the closed-loop control with a Model Predictive Controller (MPC) as seen in Fig. \ref{system}.

For methods that are not learning a policy online, as most Reinforcement Learning techniques or end-to-end MPC variants \cite{Amos:2018:DME:3327757.3327922}, the resulting controller behavior primarily depends on the accuracy and expressive capacity of the dynamics model. The idea of choosing the appropriate quantities to describe a task is not new. However, it tends not to be particularly highlighted when the focus is a simulated task where any control quantity can be readily available. In real systems, the available quantities are constrained by hardware and perception capabilities, and the method needs to reflect that.


In this paper, we present a velocity-resolved formulation for contact-rich tasks and propose a prediction method for food-cutting dynamics that can be used in model-based control schemes.  The main contributions of this work are summarized below:
\begin{enumerate}
    \item In contrast to our previous work \cite{mitsioni2019data}, the learning problem is reformulated to explicitly incorporate information related to the complicated dynamics of the task and how they are affected by the robot’s cutting actions.
    \item The proposed training scheme for multi-step prediction gradually tackles the difficulty of long-term predictions, rather than hand-picking layers to initialize at different stages of training as in \cite{deepmpc,mitsioni2019data}
    \item Our modelling and training approach can provide models that have consistently good performance and exhibit a fine-grained understanding of the task dynamics when performing within an MPC.
\end{enumerate}

\section{Related Work}\label{relatedwork}
Robotic cutting has been treated in a multitude of ways, primarily based on more traditional control approaches. In \cite{Jung1999} the authors employed an impedance controller with adaptive force tracking for a simulated object with non-homogeneous stiffness. An adaptive position controller with gradient-based estimation of the desired force was presented in \cite{Zeng2002} with simulated results as well. More recently, \cite{Long2014} proposed force control combined with visual servoing to adjust the tracked trajectories for the task of cutting deformable soft objects, while minimizing the required cutting force. A combined hybrid force/position control approach was presented in \cite{Mu2019} to cut two classes of non-deformable objects. These works all introduce physics-based mechanics that lead to a well-defined problem. However, their applicability is limited as they would require additional computations to be applied to a larger variety of cases. 

Data-driven approaches can address this issue by approximating the interaction dynamics, resulting in a single model that is capable of treating several object classes. A method that employs deep networks to approximate the dynamics for this task was first introduced in \cite{deepmpc}. Although the method's performance was evaluated on an extensive dataset, the generalization ability to unseen classes was not examined. Additionally, the proposed network outperformed several baselines but it was not clear whether there was need for the complex architecture and training procedure, as only one architecture was evaluated within the MPC. This approach was revisited by our group in \cite{mitsioni2019data} after being reformulated into a velocity resolved control problem, but with the same underlying network structure and training procedure. Unseen classes were included in the evaluation but there was no further examination of the network and its training.

In contrast to this work, our work \cite{mitsioni2019data} was an investigation of how a contact-rich task, associated with torque information, can be adapted to a different hardware setup and the constraints this imposes. Embedded in its modelling choices, was the assumption that the dynamics of cutting  can be described by a nonlinear mapping of the form $\textbf{s}_{t+1} = F(\textbf{s}_t)$. Part of what we considered as state (forces, displacements), reflected the intended action through the external forces measured by the sensor. By choosing this model, the focus shifts on the response of the \textit{coupled} object-knife system which has already incorporated the effect of the control input and thus, does not offer a proper formulation for a dynamics learning scheme. Furthermore, this hypothesis effectively considers the potentially delayed effect of the control input negligible, which is not always a valid assumption, especially in a nested control scheme, where there is no guarantee that the desired force will be reached. Notably, in \cite{Sharma2019} the authors have worked on the same task but their focus is how to learn a semantic representation rather than the dynamics.

Despite the limited amount of works for this particular application, data-driven methods in contact-rich scenarios have shown promising results. Demonstration-based methods \cite{billard1, Huang2016, deformable, peginhole}, are well suited due to their sample complexity, but infeasible for cutting with force feedback as there is no practical way to distinguish the demonstrator's exerted wrench from the object's. Reinforcement Learning, when actively focusing on sampling complexity, is a competitive alternative for real-world, contact-rich tasks. Recently, in \cite{Yfan} the authors proposed an actor-critic that is guided by supervised learning to account for sample complexity and safety, but still required 1.5 hours for an assembly task that has a smaller range of dynamics than cutting. Another method that reduces the sample complexity was presented in \cite{Johannink2018ResidualRL}. The authors actively leveraged a hand-engineered controller as a basis for a policy they optimize online, thus splitting the problem into a trajectory tracker and an adaptive corrective behavior. Their method greatly reduced the sample requirements through sim-to-real transfer but still depended on a simulated environment and unsupervised exploration, neither of which are available for our task.

A central part of the suggested training approach is curriculum training  \cite{Bengio:2009:CL:1553374.1553380} which we combine with learning rate decay to avoid prediction error accumulation and facilitate training. Curriculum training has been applied in several different contexts but to the best of our knowledge, not as a horizon curriculum for multistep prediction. An exception is the work in \cite{Ebert2018} where it is used for image registration and the authors gradually increase the temporal distance between the images. Other applications of curriculum training include mini-batch frequency selection \cite{Xu2017}, sequence prediction in natural language processing \cite{DBLP:journals/corr/RanzatoCAZ15}, equation learning \cite{maxplank3} and finally, encoding positions and velocities from pixels in simulated control tasks \cite{Jonschkowski2017}.



\section{Problem Formulation}
Consider a robotic manipulator equipped with force sensing. Let $\mathbf{p} \in \mathbb{R}^3 $ denote the translation part of the end-effector pose in the world frame and $\mathbf{f}_s \in \mathbb{R}^3 $ the force measurements. Let further  $\mathbf{p}_d,  \dot{\mathbf{p}}_d$ denote the end-effector's desired position and velocity, $\mathbf{f}_r$ the reference force and $\mathbf{u}$ a velocity control input. In order to follow a predefined trajectory in a compliant manner, we can employ a variant \cite{Siciliano:2007:SHR:1209344} of velocity-resolved (inverse) damping control, 
\begin{equation}\label{general_control}
    \mathbf{u} = \mathbf{K}_a(\mathbf{f}_s - \mathbf{f}_r).
\end{equation}
We can then define the desired compliant behavior as 
\begin{equation}\label{damping_behavior}
    \mathbf{f}_r =  \mathbf{K}_a^{-1}(\mathbf{K}_p \mathbf{e} -  \dot{\mathbf{p}}_d)
\end{equation}
where $\mathbf{K}_p,\, \mathbf{K}_a \in \mathbb{R}^{3\times 3} $ are the stiffness and compliance gain matrices and $\mathbf{e} = \mathbf{p}-\mathbf{p}_d$ the position error.

Substituting Eq.~\eqref{damping_behavior} in Eq.~\eqref{general_control} and noting that the control input corresponds to the end-effector's Cartesian velocity, results in the desired dynamic behavior
\begin{equation}\label{admittance_law}
    \dot{\mathbf{e}} + \mathbf{K}_p \mathbf{e} = \mathbf{K}_a \mathbf{f}_s
\end{equation}
where $\dot{\mathbf{e}}=\dot{\mathbf{p}} -  \dot{\mathbf{p}}_d \,\in \mathbb{R}^3$ is the velocity error.

To plan a cutting trial, it is necessary to define a desired trajectory $\mathbf{p}_d,\, \dot{\mathbf{p}}_d$ and choose appropriate controller gains $\mathbf{K}_a, \mathbf{K}_p$ with regard to the object class. A controller capable of handling the variation in the contact properties would require variable stiffness gains. However, it would still be impossible to design a single, fixed trajectory that addresses the non-homogeneous sizes of the objects. Alternatively, we can model the dynamics of the contact as a discrete-time dynamics function $\hat{\mathbf{p}}_{t+1} = f(\mathbf{p}_t, {\mathbf{f}_s}_t, {\mathbf{f}_{r}}_t)$ and determine an optimal reference force such that it minimizes a cost  $C(\mathbf{p}_t ,{\mathbf{f}_{r}}_t)$ (see Appendix) over a time horizon $T$, by solving the optimization problem
\begin{equation}\label{optimal_control}
\begin{split}
\mathbf{f}_{r}^* & = \argMin_{\mathbf{f}_r} \sum_{k=0}^{T} C\big(\hat{\mathbf{p}}_{t+k}, {\mathbf{f}_{r}}_{t+k}\big). \\
\end{split}
\end{equation}

In this work, we parametrize the dynamics function $f(\mathbf{p}_t, {\mathbf{f}_s}_t, {\mathbf{f}_{r}}_t)$ as a deep network that receives current positions, measured and reference forces, and outputs the estimated future positions. We define the model's state as the augmented state vector $\mathbf{{x}}_t = [ {\mathbf{p}}_t^T,{\mathbf{{f}}_{s}}_t^T]$ and denote $\mathbf{v}_t = {\mathbf{f}_{r}}_t$, resulting in the formulation
\begin{equation}\label{dynamics}
    \mathbf{\hat{p}}_{t+1} = f(\mathbf{x}_t, \mathbf{v}_t).
\end{equation}

This network is then used in conjunction with an MPC to determine the optimal reference force in Eq. \eqref{optimal_control}.  In contrast to our earlier work \cite{mitsioni2019data}, we model the effect of the control input explicitly through the reference force and decouple it from the interaction one, as discussed in  Section \ref{relatedwork}. Expressing the reference force as a function of the desired velocity and the position error, allows to encompass the possible delayed effects of the control input as we instill information about the divergence from the desired trajectory due to friction. This results in a more clear and concise formulation that offers a better representation for the learning task. 

To further demonstrate the importance of considering $\mathbf{f}_{r}$, we can transform the initial data space that includes multi-step sequences of 6 or 9 features, into a 2-D one and visualize the data with t-SNE \cite{maaten2008visualizing}. t-SNE is a probabilistic dimensionality reduction technique that projects data into their low-dimensional embedding in a non-linear way, while trying to preserve their probabilistic distribution. To have a fair comparison, we used the same dataset\footnote{For all the visualizations, the dataset consists of 24 different cutting trials for 6 different objects. The t-SNE hyperparameter "perplexity" was set to 30 and the max number of iterations to 3000.}
 but omitted the $\mathbf{f}_{r}$ inputs for the latter case. The two resulting datasets went through the same pre-processing as in Eq. \eqref{block equation} and \eqref{relative displacement}. 

\begin{figure}[t]
\centering
\begin{subfigure}{0.5\linewidth}
  \centering
  \includegraphics[width= \linewidth]{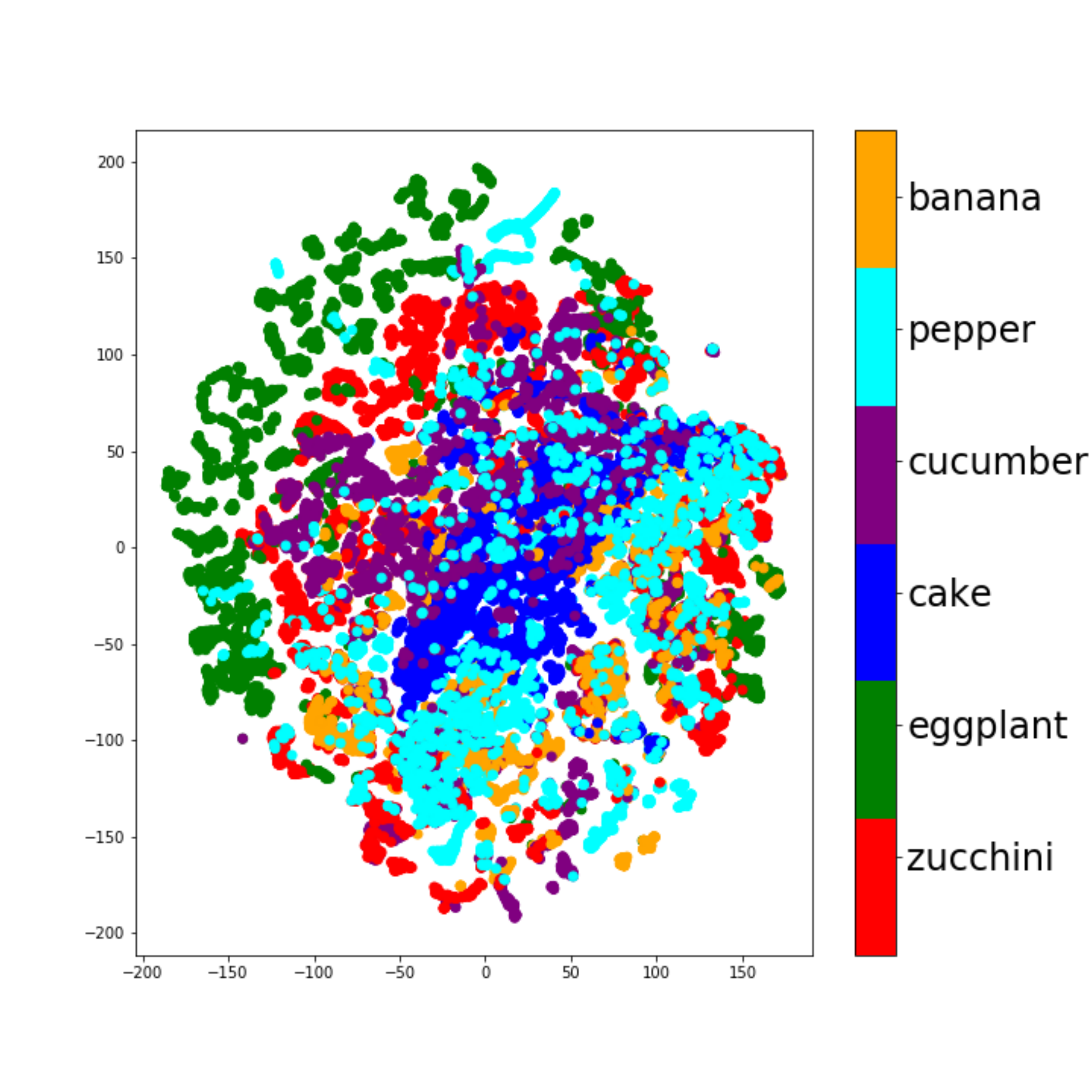}
  \caption{\footnotesize t-SNE visualization of $\mathcal{D}_1$}
  \label{old}
\end{subfigure}%
\begin{subfigure}{0.5\linewidth}
  \centering
  \includegraphics[width = \linewidth]{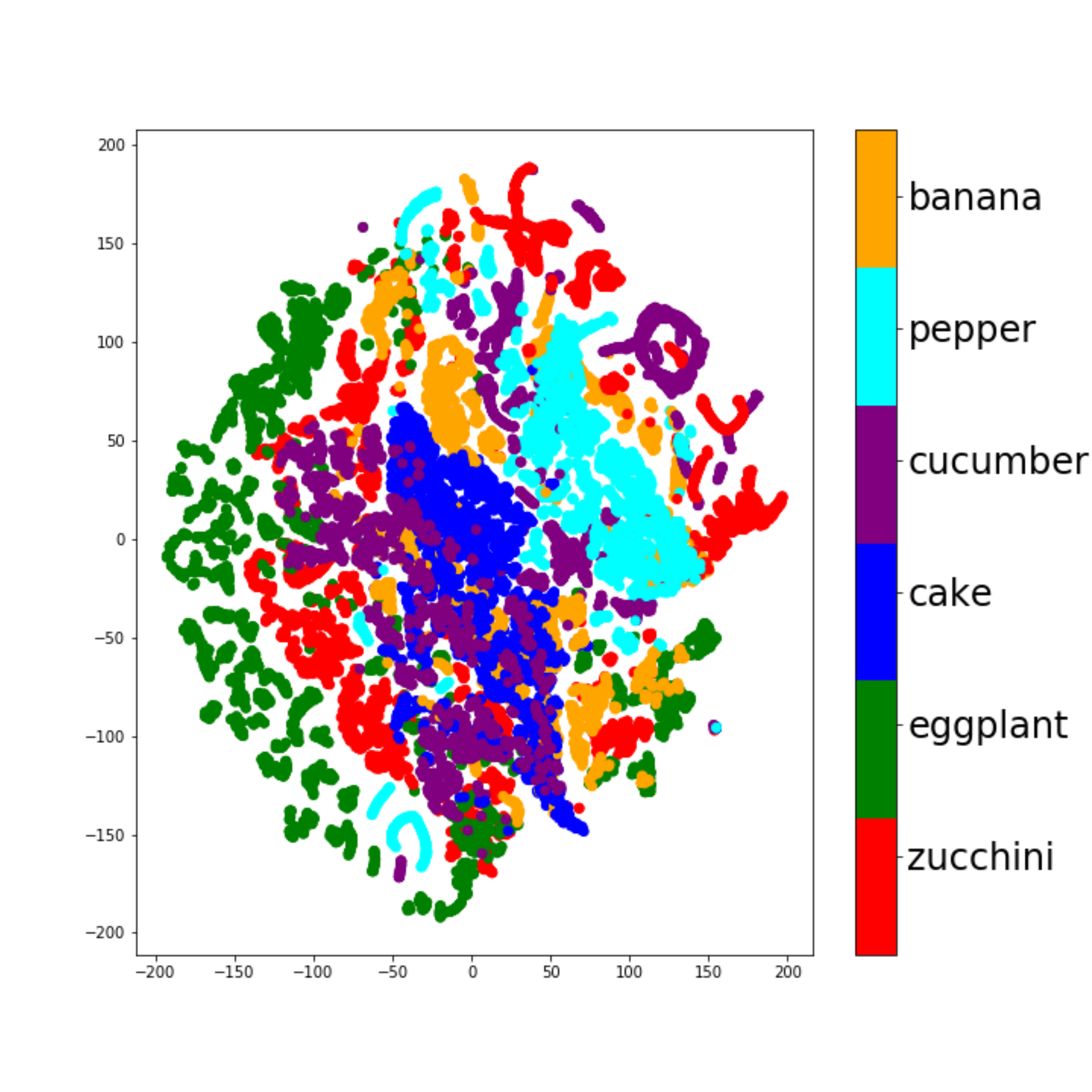} 

  \caption{\footnotesize t-SNE visualization of $\mathcal{D}_2$}
  \label{new}
\end{subfigure}
\caption{\footnotesize Embedding space of the two datasets. Adding $\mathbf{f}_r$ to the learning process produces more coherent clusters suggesting a more informative latent space.}
\label{original_data}
\vspace{-0.5cm}
\end{figure}

Fig. \ref{old} shows the results of the dataset $\mathcal{D}_{1} = \{\mathbf{p}, \mathbf{f}_s\}$ and correspondingly, Fig. \ref{new} the ones from the dataset we propose for this task, namely $\mathcal{D}_{2} = \{ \mathbf{p}, \mathbf{f}_s, \mathbf{f}_{r}\}$.
With dataset $\mathcal{D}_1$ , as seen in Fig. \ref{old}, there is no specific structure in the embedding except for the eggplant class as the class dynamics are the most easily distinguishable during the task due to the object's texture.
In comparison, adding $\mathbf{f}_{r}$ and visualizing $\mathcal{D}_2$, produces more coherent clusters. The central part of the plot is mostly occupied by easier to cut classes and as we are moving peripherally outwards, we get cases of stiffer materials. Although we are not interested in classifying the objects, a more cohesive embedding indicates that $\mathcal{D}_2$ is a more informative representation and henceforth, the networks we compare in the experimental section are trained on these features.

\section{Method}
\subsection{Modelling Cutting}\label{modelling}
Modelling cutting analytically is a complicated process due to its frictional properties as well as the separation of fibers \cite{Cotin2000}. Nevertheless, it can be approximated given appropriate inputs and a model that is adequately expressive to capture the nonlinear temporal and spatial variations.

In the context of this work, we are interested in representing the interaction dynamics between the manipulator and the object as the transition function in Eq.~\eqref{dynamics}. Therefore, the dataset needs to reflect the current state of the system and the delayed effect of the controller's input. Traditionally, to characterize the dynamics during an interaction task, the terms of mechanical impedance and admittance are introduced, which are characterized as mappings between velocities and forces. However, velocities are usually noisy and not easy to learn from. In addition, considering joint velocities unnecessarily increases the difficulty of the task as the approximator is implicitly required to learn the robot's kinematics. 

Instead, we employ relative displacements over time to approximate a generalized notion of velocity, similar to \cite{deepmpc, mitsioni2019data}. To achieve that, the input features for the learning module are not treated as single time-steps but form non-overlapping blocks of sequences. Block $b$ of length $M$ is then given by 
\begin{equation}\label{block equation}
  \mathbf{X}^M_b = [\mathbf{x}^T_{bM} ; \, \ldots \,;\, \mathbf{x}^T_{(b+1)M-1}], \quad\in \mathbb{R}^{M\times6}.
\end{equation}
If we denote the positional elements of $\mathbf{X}^M_b$ as $\mathbf{P}^M_b$ and an all-ones vector of length $M$ as $\mathbf{1}^M$, the transformation from positions to relative displacements is done by subtracting the past block's last position from every position in the current one
\begin{equation}\label{relative displacement}
\mathbf{\Delta \mathbf{p}}^M_b = \mathbf{P}^M_b -{ \mathbf{1}^M} \mathbf{p}_{(b-1)M-1}.
\end{equation}

Dropping the superscript $M$ for brevity, the network's input is then $\mathbf{X}_b = [\mathbf{\Delta \mathbf{p}}_b, {\mathbf{f}_s}_b]$. Through this transformation, we also ensure that the network will not overfit to absolute positions, which do not carry the same amount of information as they depend on the object's size. Since we are using relative displacements and sample every $5 \textrm{ms}$, the magnitude of the positional part is significantly smaller than the remaining features of the input vector. To ensure consistency in the input range, we normalize the features to zero mean and unit standard deviation. 



\subsection{Network Architecture and Training}\label{network_training}
In this work, we chose to employ an LSTM network as opposed to Recurrent Neural Networks (RNN) used in \cite{deepmpc, mitsioni2019data}. While more complex than a regular RNN, LSTMs have proven to be suitable for learning sequences and dependencies further in time \cite{Chung2014EmpiricalEO}, which is appealing for a task that requires modelling of temporally and spatially varying dynamics. 

Although Eq.~\eqref{dynamics} is referring to one-step predictions, predicting the positions at a time-step $H$ ahead into the future can be achieved by recursively using the intermediate prediction $\hat{\mathbf{p}}_{b+i}, \, i \in [1, H/M]$ as inputs until we reach the desired horizon i.e.
\begin{align*}
    \hat{\mathbf{p}}_{b+1} &= f(\mathbf{X}_b, \mathbf{v}_b) \\
    \hat{\mathbf{p}}_{b+2} &= f(\hat{\mathbf{X}}_{b+1}, \mathbf{v}_{b+1}) \\
    \dots \\ 
    \hat{\mathbf{p}}_{b+H} &= f(\hat{\mathbf{X}}_{b+H-1}, \mathbf{v}_{b+H-1})
\end{align*}{}
 This results in a sequence-to-sequence prediction that, for a robot working at $200 \textrm{Hz}$ and with $M = 10$, instead of predicting to $t_H = 0.15 \textrm{s}$,  the new horizon will be $b_H = 0.15/0.05 = 3$ blocks ahead.

A common problem with the recursive approach is error accumulation, since predictions are used in place of observations. To avoid that, we propose to train the system with a curriculum strategy that gradually increases the difficulty of the prediction goal. Practically, this amounts to progressively predicting further ahead in the future by increasing the horizon. However, the abrupt difference in difficulty might lead the system into instability, or it might render the hyperparameters used for the easier problem unsuitable. Therefore, we apply learning rate decay when the horizon changes, so that learning is adjusted to the new horizon smoothly and the gradient steps are affected less by the change, especially during the transitions.

\subsection{Model Predictive Control}\label{MPC}
We treat the problem in Eq.~\eqref{optimal_control} with an MPC \cite{RHC} that instead of solving the optimization problem for an infinite horizon, executes the first step of the solution and then re-samples the current state. By doing so, it alleviates the need for a global, open-loop, plan that would require model-plant mismatches or abrupt discontinuities to be treated a priori. Instead, receding horizon controllers correct them by sampling the real system state at the next optimization round. 

We manage the compliant reaction to the environment separately as seen in Fig. \ref{system}, and use $\mathbf{f}_{r}$ as a feature for the dynamics model and the optimization variable of Eq. \eqref{optimal_control}.
Finally, for the MPC state we do not consider the full pose of the end-effector, but simplify the problem by only treating the translational parts of the cutting motion (axes $Y,\, Z$ in Fig. \ref{system}). The motion on the remaining axes is controlled through a set-point stiffness controller. 

\section{Evaluation}
 For all of the following experiments, the training set, or seen classes, includes trials for 6 object classes, $\mathcal{D}_{seen} = \{banana, pepper, cucumber, cake, eggplant, zucchini\}$. In Sections \ref{forward_predictions} and \ref{online_experiments} we also examine the generalization ability of the different models by adding 3 completely unseen classes, $\mathcal{D}_{unseen} = \{cheese, potato, lemon\}$.  For the experiments, the blocks are formed by $M = 10$ time-steps which correspond to $0.05s$ of measurements and in Sections \ref{forward_predictions}, \ref{online_experiments} the prediction horizon is set to $H = 5$ blocks. More information regarding the models' parameter values and experimental details are listed in the Appendix.

\subsection{Data Collection}\label{data_collection}
We collected data from cutting trials executed with the controller in Eq. \eqref{general_control}, where the desired behavior in Eq. \eqref{damping_behavior} was commanded as a trajectory ($\mathbf{p}_d, \dot{\mathbf{p}}_d$). The gains, as well as the sawing rate for the desired trajectory, were tuned depending on the object class in order to record a large variety of interaction modes. We included interactions with optimal parameters for the specific object, interactions with appropriate parameters for the whole class, as well as parameters that could accommodate all the classes, albeit not in an optimal manner. All the cutting trials were initialized above the object, depending on its size, as to include enough samples of free-space motion.

\begin{figure}[t]
    \centering
    \includegraphics[width = 0.8\linewidth]{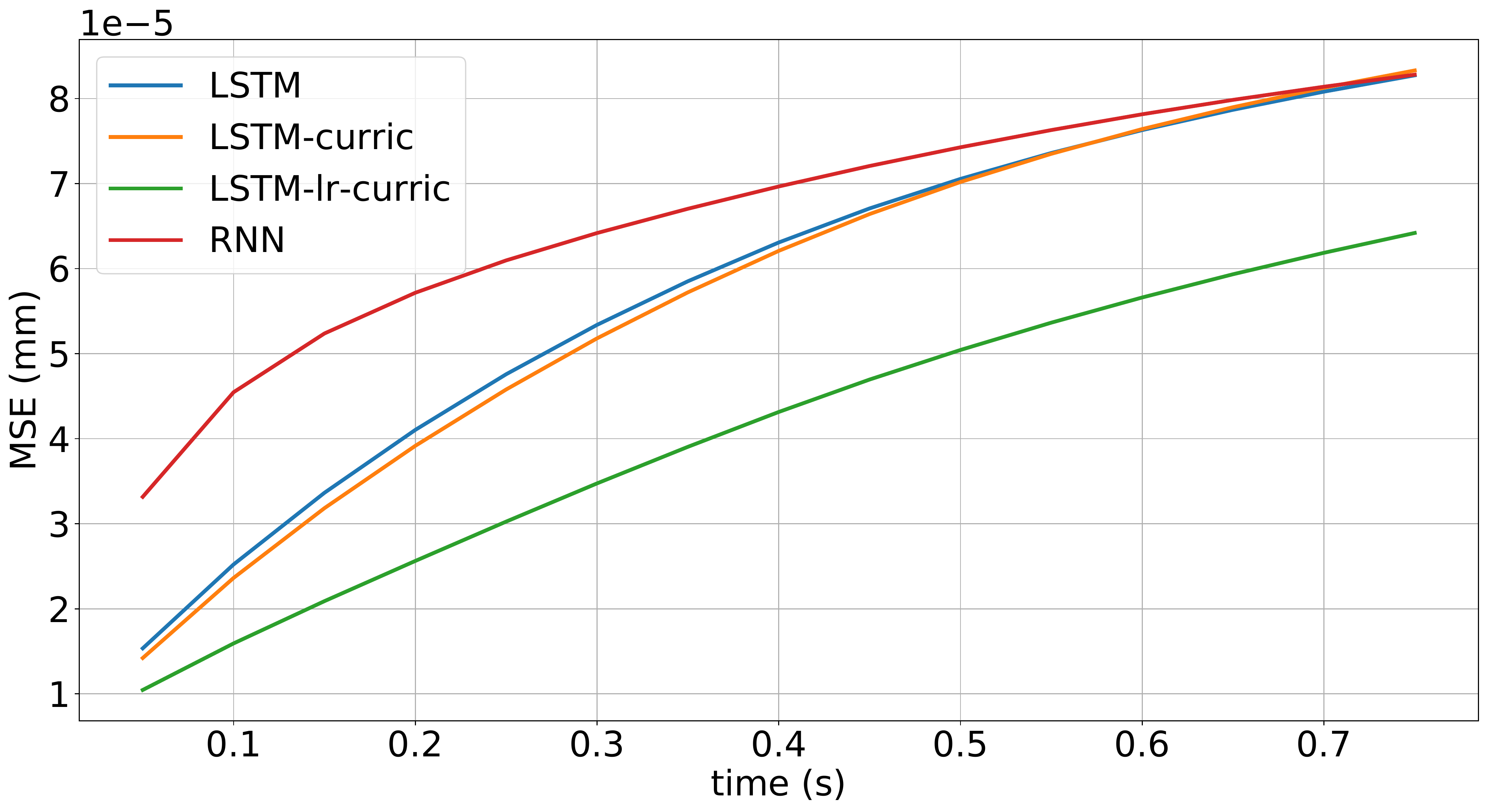}

    \caption{\footnotesize Evolution of the Mean Squared Error for a total horizon of 0.75 secs which is equivalent to predicting 15 blocks ahead in the future. The predictions and the groundtruth are expressed as relative displacements.}
    \label{forward_preds_plot}
    \vspace{-0.3cm}
\end{figure}

\begin{table*}[ht]
\centering
\resizebox{0.96\textwidth}{!}{%
\begin{tabular}{c  c c c c c c c c c | c}
\hline
 & \textbf{Cake} & \textbf{Zucchini} & \textbf{Cucumber} & \textbf{Banana} & \textbf{Pepper} & \textbf{Eggplant} & \textbf{Cheese*} & \textbf{Potato*} & \textbf{Lemon*} & \textit{\textbf{Avg.}} \\ \hline
\textbf{\texttt{RNN}} & 1.08 & 1.04 & 1.49 & 1.59 & 0.92 & 1.42 & \textbf{0.71} & 0.55 & N/A & \cellcolor[HTML]{EFEFEF}\textit{1.13**} \\ 
\textbf{\texttt{LSTM}} & \textbf{1.07} & 2.34 & 0.96 & 1.72 & 2.08 & 2.11 & 1.69 & 0.68 & 1.43 & \cellcolor[HTML]{EFEFEF}\textit{1.56} \\ 
\textbf{\texttt{LSTM-c}} & 1.51 & 1.41 & 1.68 & \textbf{1.19} & 1.21 & 0.98 & 2.18 & \textbf{0.51} & \textbf{0.99} & \cellcolor[HTML]{EFEFEF}\textit{1.29} \\ 
\textbf{\texttt{LSTM-lr-c}} & 1.26 & \textbf{0.91} & \textbf{0.71} & 2.31 & \textbf{0.72} & \textbf{0.95} & 0.73 & 1.76 & 1.80 & \cellcolor[HTML]{EFEFEF}\textit{1.24} \\ \hline
 \textit{\textbf{Avg.}} & \cellcolor[HTML]{EFEFEF}\textit{1.23} & \cellcolor[HTML]{EFEFEF}\textit{1.42} & \cellcolor[HTML]{EFEFEF}\textit{1.21} & \cellcolor[HTML]{EFEFEF}\textit{1.71} & \cellcolor[HTML]{EFEFEF}\textit{1.23} & \cellcolor[HTML]{EFEFEF}\textit{1.36} & \cellcolor[HTML]{EFEFEF}\textit{1.33} & \cellcolor[HTML]{EFEFEF}\textit{0.88} & \cellcolor[HTML]{EFEFEF}\textit{1.16**} & \cellcolor[HTML]{EFEFEF} \\ \hline
\end{tabular}%
}
\caption{\footnotesize Mean cost across trials. Object classes denoted with an asterisk belong to $\mathcal{D}_{unseen}$. The results denoted with a double asterisk do not take into account the failed attempts. }
\label{MPC_results_costs}
\vspace{-0.5cm}
\end{table*}

\subsection{Prediction Performance}\label{forward_predictions}
In this section, we investigate the effects of the proposed training approach on the prediction performance of the networks modelling the dynamics. To evalue this , we compare an RNN architecture (\texttt{RNN}), a baseline LSTM network trained directly for 5-block prediction (\texttt{LSTM}), an LSTM trained with horizon curriculum (\texttt{LSTM-c}) and finally the LSTM trained with the proposed combination of horizon curriculum and decaying learning rate (\texttt{LSTM-lr-c}). The \texttt{RNN} was structured and trained with the 3-stage approach in \cite{mitsioni2019data}. In the experiments, we investigate whether a simpler architecture can capture the dynamics, the effect of the proposed training approach and finally how these changes can affect the generalization ability over different object classes.

Firstly, we examine the evolution of the mean L2 error between predicted and groundtruth trajectories as the prediction horizon increases up to $0.75s$ (or 15 blocks) into the future. Note that the trajectories consist of relative displacements, hence their magnitude. For this experiment, the networks were trained on a dataset containing 34 cutting trials over 6 objects (210564 data points in total), while the validation set includes 15 independent trials over the same object categories (93447 data points in total).

From the results in Fig. \ref{forward_preds_plot} it can be seen that for short horizons, the LSTM networks have comparable results, while the \texttt{RNN} displays a much higher error. As the horizon increases, the performance of all the networks, except \texttt{LSTM-lr-c}, degrades to the same point. Before the prediction horizon reaches $t = 0.45s$, \texttt{LSTM-c} has only marginally better results than its simpler counterpart, showcasing that simply employing a learning curriculum is not enough to boost the predictive performance. Finally, throughout the experiment, \texttt{LSTM-lr-c} significantly outperforms all of the baselines, supporting that the combination of learning rate decay and curriculum training results in better performance that scales well with the prediction horizon.

Secondly, we report the average MSE during forward predictions on a test set for a prediction horizon of $H = 5$ blocks. For this purpose, we performed trials for 5 object classes that were also in the training set $\mathcal{D}_{seen}$ and the additional 3 classes in $\mathcal{D}_{unseen}$. We recorded two repetitions for three different values of $\mathbf{K}_a$ amounting to a total of 30 trials with seen classes and 18 trials with unseen ones. 

Table \ref{test_set} shows the corresponding results for each model on seen and unseen classes, as well as the total MSE for both cases. It is evident that \texttt{LSTM-lr-c} is consistently better than the rest of the models and generalizes well to the unseen cases. It is interesting to observe that despite its poor scaling as the horizon grows, the \texttt{RNN} model shows slightly better results than the LSTM baselines in both datasets. This reinforces the results from the previous section concerning the training procedure and further indicates the usefulness of combining curriculum training with learning rate decay.

\begin{table}[h]
\resizebox{0.9\linewidth}{!}{%
\begin{tabular}{c c c c}
\hline
\textbf{Model} & \textbf{\begin{tabular}[c]{@{}c@{}}Seen classes\\ \tiny ($10^{-5}$ mm)\end{tabular}} & \textbf{\begin{tabular}[c]{@{}c@{}}Unseen classes \\ \tiny ($10^{-5}$ mm)\end{tabular}} & \textbf{\begin{tabular}[c]{@{}c@{}}Total \\ \tiny ($10^{-5}$ mm)\end{tabular}} \\ \hline
\textbf{\texttt{RNN}}            & 2.08                       & 3.33                                                    & 2.55                  \\ 
\textbf{\texttt{LSTM}   }        & 2.26                       & 3.75                                                      & 2.82                 \\ 
\textbf{\texttt{LSTM-c} }   & 2.30                       & 3.94                                                       & 2.92                  \\ 
\textbf{\texttt{LSTM-lr-c}} & \textbf{1.37}                        & \textbf{2.29 }                                                      & \textbf{1.72 }                 \\ \hline
\end{tabular}%
}
\caption{\footnotesize Test performance. "Seen classes" include unseen datasets but on objects that have been treated in training as opposed to the "Unseen Classes" that have never been encountered.}
\label{test_set}
\vspace{-0.4cm}
\end{table}


\subsection{Robotic experiments}\label{online_experiments}

\begin{figure*}[t]
\centering
\begin{subfigure}[b]{0.38\textwidth}
  \centering
  \includegraphics[width = \linewidth]{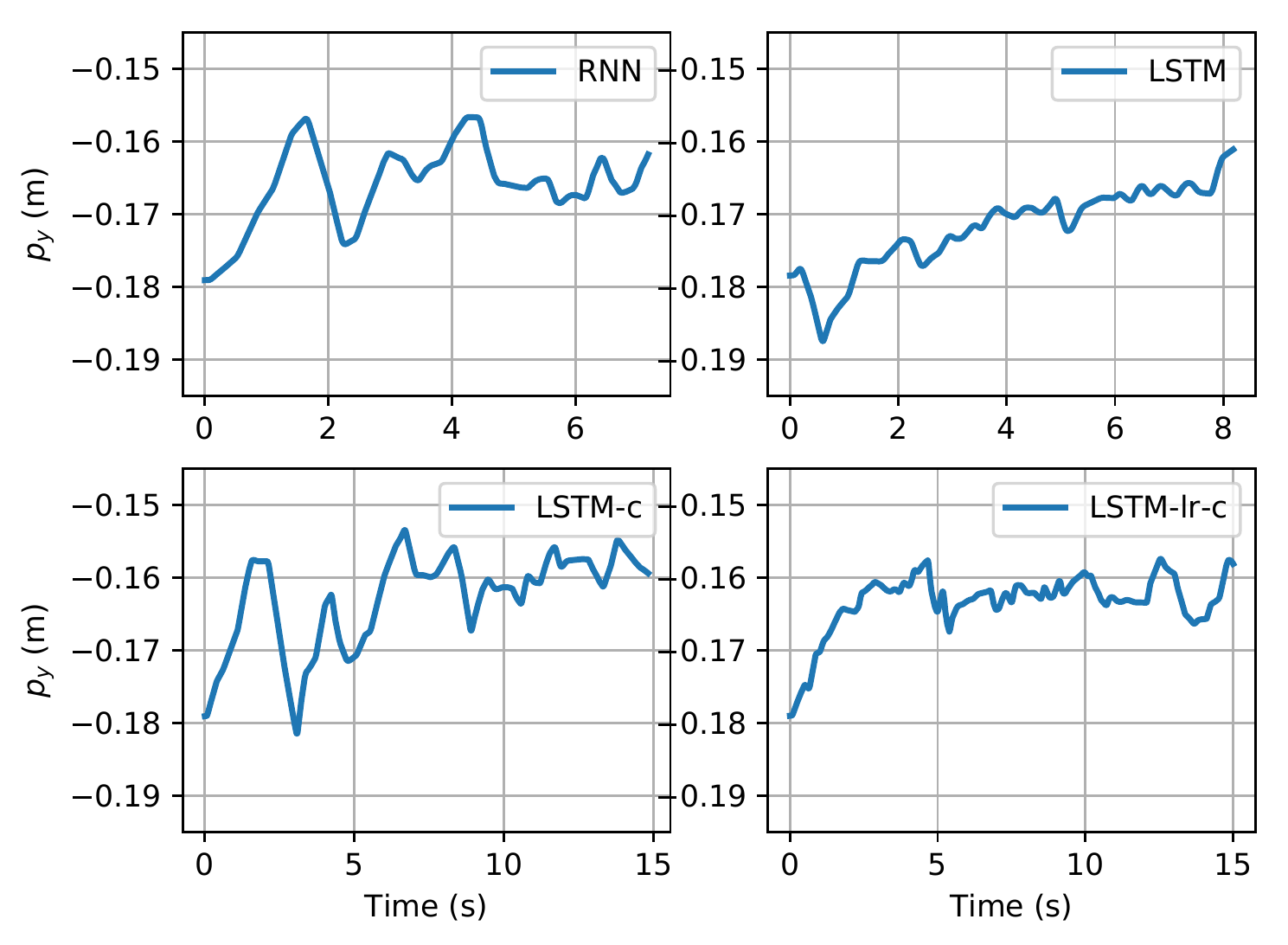}
    \caption{\footnotesize Sawing motion for the four model baselines on a soft object.}
    \label{cutting_online_plot_cake}
\end{subfigure}%
  \vspace{-0.05cm}
\begin{subfigure}[b]{0.38\textwidth}
  \centering
    \includegraphics[width = \linewidth]{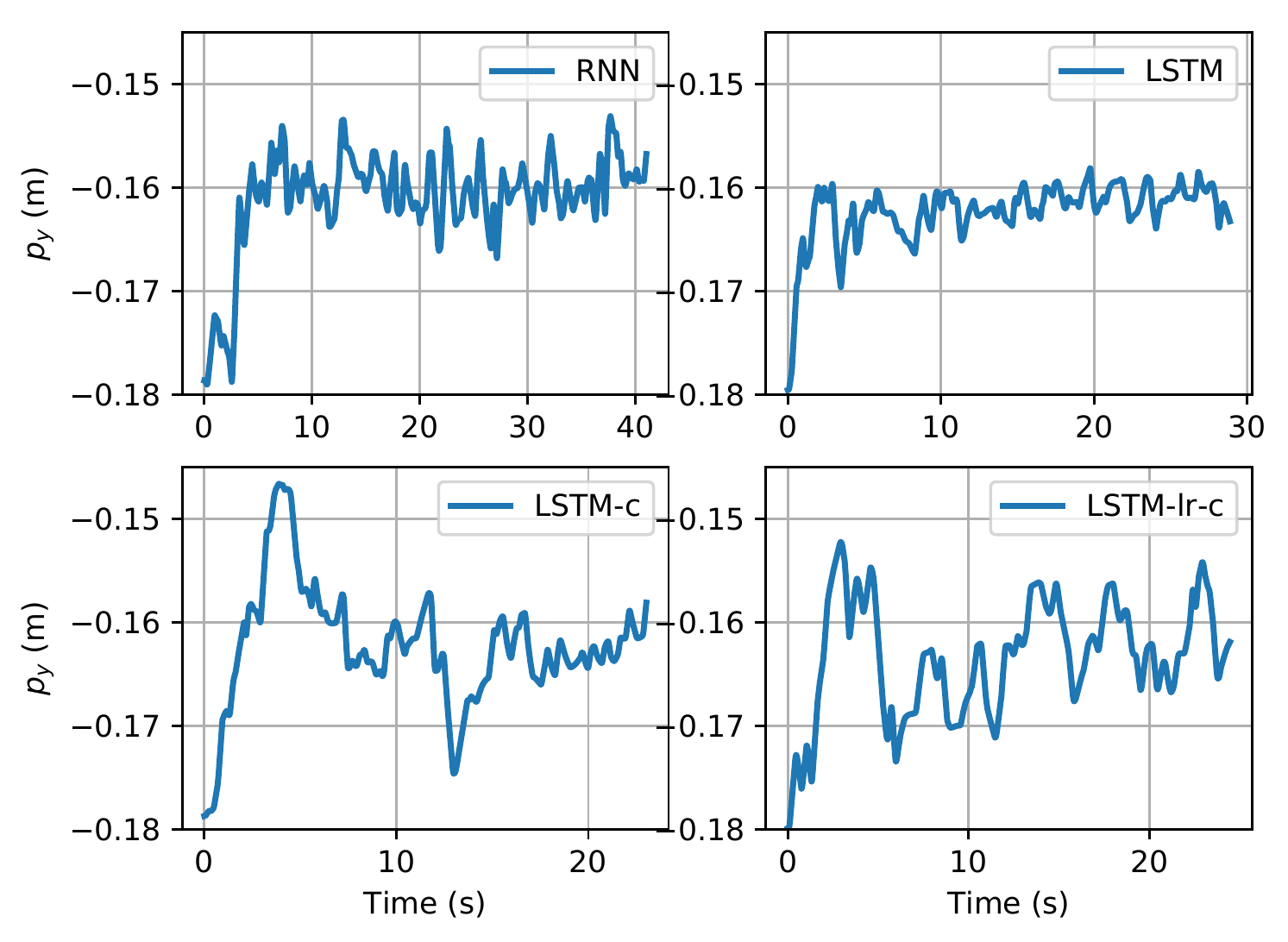}

    \caption{\footnotesize Sawing motion for the four model baselines on a stiff object. }
    \label{cutting_online_plot_eggplant}
\end{subfigure}
\caption{\footnotesize The sawing trajectories followed by the controller during two very distinct cutting cases. \texttt{LSTM-lr-c} leads the controller to insightful strategies depending on the dynamics of the class.}
\label{cutting_trajz}
\vspace{-0.3cm}
\end{figure*}


Even though deep networks can efficiently model non-linear mappings, when training a network for a dynamics model, the focus should be on the closed-loop behavior. A good prediction accuracy is a good indication of the modelling capabilities but does not necessarily reflect what that behavior will be as both training and testing are done on trajectories executed by a trajectory-tracking controller, which is different than the cost-based MPC during online deployment. The desirable properties of that closed-loop system are primarily qualitative and difficult to express quantitatively. Considering that we are aiming to construct an intelligent system that is handling the task of food-cutting generally, it is of paramount importance that the results for different models are consistently good and can tackle different object types. In that light, failure to complete trials for a given object type, such as the \texttt{RNN} model in Table \ref{MPC_results_costs}, should be weighted more than good training and validation results.

To evaluate the models' performance within the controller, we executed a series of experiments with the 9 different object classes in $\mathcal{D}_{seen}$ and $\mathcal{D}_{unseen}$. In the experiments, we used a YuMi-IRB 14000 collaborative robot with an OptoForce 6axis Force/Torque sensor mounted on its wrist. For every model, we executed 5 trials per class. Throughout the trials, we set $\mathbf{K}_a = 0.003\mathbf{I}_3$ and kept the same cost function as reported in the Appendix. The trial ended successfully only when the knife had reached the cutting board. In any other case, e.g. if the execution time exceeded a minute, the trial was considered unsuccessful and the results were discarded. Since we have included sequences of free-space motion in the training data, we did not initialize the trials with the knife already in the object as in \cite{deepmpc, mitsioni2019data}, but directly above it with no further indications of the object's location.

It should be noted that due to the robot's hardware limitations, stiffer objects often caused the torque limit to be surpassed, leading the robot to shut down, which constituted a failure and the trial was repeated. This was especially evident while evaluating the \texttt{RNN} architecture on lemons as the closed-loop behavior did not exhibit the necessary sawing motion to break friction. This hampered the downwards progress, or simply caused a hardware failure, making it impossible to collect successful trials. For this reason, in Table \ref{MPC_results_costs} the associated results are marked with a double asterisk to denote that \texttt{RNN} trials were omitted when calculating the average scores.

The only modifications made to the objects were size adjustments if their width exceeded the knife's length. We did not alter the objects' height as that could alter the dynamics of certain classes. Additionally, because of the aforementioned hardware limitations, when treating eggplants and lemons, we created a small slit on the object's surface to decrease the exerted torque during the initial contact and avoid reaching the torque limits before the controller had time to employ sawing motions.

Since most of the objects did not have homogeneous shapes, the required cutting length was often not the same. Consequently, the accumulated costs per trial would be a misleading and an incomparable metric as longer trajectories do not necessarily signify worse performance. Instead, in Table \ref{MPC_results_costs} we assess the online performance through the mean cost achieved by the models. For almost half of the classes, \texttt{LSTM-lr-c} again outperforms the rest of the baselines and achieves better mean costs. However, it is interesting to notice that \texttt{LSTM-c}, despite having the highest MSE during forward predictions, still manages to perform well, and more importantly, accomplishes the best scores in 2 out 3 unseen cases. Finally, even though \texttt{RNN} failed to complete a cut on lemons, it still has notable performance on several classes. This is partially due to the fact that the strategies it resulted in revolved mostly around slicing the object instead of sawing. However, because this behavior was more aggressive, it often led to failed trials because of the hardware limits. 

In a successful cutting trial, it is straightforward to surmise that the main objective is downwards motion. Nevertheless, the sawing motion is related to the downward progress, as it enables it by breaking friction and minimizing the sheer force otherwise required. Consequently, apart from the mean cost, a crucial point of evaluation for the dynamics models is whether they lead the controller to infer useful strategies for each object class. For objects that are stiffer, fine-grained understanding of the dynamics should drive the strategy around sawing, while for the softer ones, it should deem it unnecessary. A qualitative demonstration of these emerging behaviors can be observed in Fig. \ref{cutting_online_plot_cake} and Fig. \ref{cutting_online_plot_eggplant} that depict the trajectories during trials on a soft (cake) and a stiff object (eggplant).

In the former case, any strategy is viable as there is no significant resistance from the material. \texttt{LSTM} that had the best cost for this class, results in minimum sawing, as it is redundant, and so is \texttt{LSTM-lr-c} despite it's worse cost-wise performance. On the other hand, \texttt{RNN}, that had almost the same cost as \texttt{LSTM}, displays similar behaviour with the worst model for this class. In the latter case of the eggplant, it is substantially more difficult to cut through the object without sawing, because of its density and firmness. \texttt{LSTM-lr-c} demonstrates the most insightful behavior with smooth sawing motions that led to the best cost. Similar behavior is adopted by \texttt{LSTM-c} that has the closest score, as opposed to \texttt{LSTM} and \texttt{RNN} that only employed low-magnitude sawing, which was not suitable for the dynamics of the class. In conclusion, even though \texttt{LSTM-lr-c} did not have the least cost for every class, it exhibited the most appropriately diverse techniques that were able to adapt efficiently to the dynamics encountered amongst the classes. 

\section{Conclusion}
In this work, we presented a data-driven framework for the contact-rich task of food-cutting. We showed that by carefully designing every step of the method, we can produce models that have consistently good predictive performance on known cases and generalize well to unseen ones. When evaluated within a predictive controller, the proposed approach achieved the best mean cost in 4 out of 9 object classes and displayed a better understanding of the dynamics as showcased by the strategies the controller adopted. In the future, it would be interesting to explore avenues that allow adaption not only on different object sizes or classes, but on completely different and more complicated cases of cutting, such as objects with a large seed. To this end, we will further investigate the design choices as to seamlessly incorporate behaviors that could be otherwise generated by switching controllers or a high-level planner.  


\section*{Appendix}\label{Appendix}
\subsection*{Network architectures}
The LSTM networks consist of a fully connected input layer of size 90 with a hyperbolic tangent activation, followed by 2 LSTM layers of hidden size 9 and a linear output layer that transforms the LSTM output to size 30. The \texttt{RNN} baseline consists of 6 fully connected layers with hyperbolic tangent activation and 2 recurrent layers with 30 units each.

\subsection*{Curriculum training}
During curriculum training, we gradually increase the horizon until we reach the desired one. For every horizon, we train the network for 10 epochs and reduce the learning rate, except for the final length prediction that we allow the network to train for 20 epochs without further changing it.  

\subsection*{Training Hyperparameters}
For all the networks we used Adam \cite{adam} with the hyperparameters learning rate (lr), weight decay (wd) and learning rate decay (gamma) set as listed in Table \ref{hyperparams}.
\begin{table}[h]
\centering
\resizebox{\linewidth}{!}{%
\begin{tabular}{|c|c|l|l|}
\hline
\textbf{Model} & \textbf{\{lr, wd, gamma\}} & \textbf{Model} & \textbf{\{lr, wd, gamma\}} \\ \hline
               & \{1e-04, 5e-04, N/A\}      &                &                            \\ \cline{2-2}
\textbf{\texttt{RNN} }           & \{1e-04, 5e-04, N/A\}      & \textbf{\texttt{LSTM-c}}         & \{1e-04, 2e-04, N/A\}      \\ \cline{2-2}
               & \{1e-04, 3e-04, N/A\}      &                &                            \\ \hline
\textbf{\texttt{LSTM}  }         & \{2e-04, 3e-04, N/A\}      & \textbf{\texttt{LSTM-lr-c}  }    & \{1e-04, 3e-04, 0.5\}      \\ \hline
\end{tabular}%
}
\caption{\footnotesize Hyperparameters used for the experimental results}
\label{hyperparams}
\vspace{-0.5cm}
\end{table}

\subsection*{Model Predictive Control}
The main components of the cost function are a term that drives the slicing motion towards the table's surface $\mathbf{p}_{table}$ and a sawing term that enables the downward progress. Since there is no fixed trajectory, the sawing term does not penalize motion within a range $d$, with an $\epsilon$ margin, around the central sawing point $\mathbf{p}_{center}$ and is quadratic beyond it. Finally, to motivate smaller-effort solutions, we include the norm of the control input. Namely, for the prediction horizon $H_b$, the cost was given by: 
\begin{align}
C(\mathbf{p}, \mathbf{u}) & = c_{\textrm{cut}}\sum_{k = 1}^{H_b}\left( \mathbf{p}^z_k - \mathbf{p}_{table}\right)^2  \nonumber
\\& + c_{\textrm{saw}}\sum_{k = 1}^{H_b}\left( \max\{0, |\mathbf{p}^y_k - \mathbf{p}_{center}| - d + \epsilon \}\right)^2  \nonumber
\\& + c_v\sum_{k = 1}^{H_b} \left\Vert  \mathbf{v}_k \right\Vert^2 \nonumber
\end{align}
where $c_{\textrm{cut}}$, $c_{\textrm{saw}}$ are positive constants weighting the contribution of the costs associated with cutting and sawing actions respectively to the total cost while $c_u$ is the weighting constant for the control input quadratic term.

To solve Eq.~\eqref{optimal_control} we use a shooting method \cite{shooting}. For every optimization iteration, we generate 25 potential inputs that act as the feasible forces for this round and choose the one associated with the lowest cost as $\mathbf{f}_{r}^*$.

\bibliographystyle{IEEEtran}
\bibliography{references}

\begin{thebibliography}{10}
\providecommand{\url}[1]{#1}
\csname url@samestyle\endcsname
\providecommand{\newblock}{\relax}
\providecommand{\bibinfo}[2]{#2}
\providecommand{\BIBentrySTDinterwordspacing}{\spaceskip=0pt\relax}
\providecommand{\BIBentryALTinterwordstretchfactor}{4}
\providecommand{\BIBentryALTinterwordspacing}{\spaceskip=\fontdimen2\font plus
\BIBentryALTinterwordstretchfactor\fontdimen3\font minus
  \fontdimen4\font\relax}
\providecommand{\BIBforeignlanguage}[2]{{%
\expandafter\ifx\csname l@#1\endcsname\relax
\typeout{** WARNING: IEEEtran.bst: No hyphenation pattern has been}%
\typeout{** loaded for the language `#1'. Using the pattern for}%
\typeout{** the default language instead.}%
\else
\language=\csname l@#1\endcsname
\fi
#2}}
\providecommand{\BIBdecl}{\relax}
\BIBdecl

\bibitem{hogan}
N.~{Hogan}, ``Impedance control: An approach to manipulation,'' in \emph{1984
  American Control Conference}, June 1984, pp. 304--313.

\bibitem{Siciliano:2000:RFC:555628}
B.~Siciliano and L.~Villani, \emph{Robot Force Control}, 1st~ed.\hskip 1em plus
  0.5em minus 0.4em\relax Norwell, MA, USA: Kluwer Academic Publishers, 2000.

\bibitem{schutter}
\BIBentryALTinterwordspacing
J.~D. Schutter and H.~V. Brussel, ``Compliant robot motion i. a formalism for
  specifying compliant motion tasks,'' \emph{The International Journal of
  Robotics Research}, vol.~7, no.~4, pp. 3--17, 1988. [Online]. Available:
  \url{https://doi.org/10.1177/027836498800700401}
\BIBentrySTDinterwordspacing

\bibitem{tune}
Y.~Karayiannidis and Z.~Doulgeri, ``An adaptive law for slope identification
  and force position regulation using motion variables,'' vol. 2006, 06 2006,
  pp. 3538 -- 3543.

\bibitem{adaptiveRev}
\BIBentryALTinterwordspacing
D.~Zhang and B.~Wei, ``{A review on model reference adaptive control of robotic
  manipulators},'' \emph{Annual Reviews in Control}, vol.~43, pp. 188--198,
  2017. [Online]. Available:
  \url{http://www.sciencedirect.com/science/article/pii/S1367578816301110}
\BIBentrySTDinterwordspacing

\bibitem{doors}
Y.~{Karayiannidis}, C.~{Smith}, F.~E.~V. {Barrientos}, P.~{\"{O}gren}, and
  D.~{Kragic}, ``An adaptive control approach for opening doors and drawers
  under uncertainties,'' \emph{IEEE Transactions on Robotics}, vol.~32, no.~1,
  pp. 161--175, Feb 2016.

\bibitem{parallel}
S.~{Chiaverini} and L.~{Sciavicco}, ``The parallel approach to force/position
  control of robotic manipulators,'' \emph{IEEE Transactions on Robotics and
  Automation}, vol.~9, no.~4, pp. 361--373, Aug 1993.

\bibitem{hybrid}
M.~H.~Raibert and J.~J.~Craig, ``Hybrid position/force control of
  manipulator,'' \emph{Journal of Dynamic Systems Measurement and Control},
  vol. 103, 12 1980.

\bibitem{Kroemer2019ARO}
O.~Kroemer, S.~Niekum, and G.~Konidaris, ``A review of robot learning for
  manipulation: Challenges, representations, and algorithms,'' \emph{ArXiv},
  vol. abs/1907.03146, 2019.

\bibitem{ATKINS2005479}
\BIBentryALTinterwordspacing
A.~Atkins and X.~Xu, ``Slicing of soft flexible solids with industrial
  applications,'' \emph{International Journal of Mechanical Sciences}, vol.~47,
  no.~4, pp. 479 -- 492, 2005, a Special Issue in Honour of Professor Stephen
  R. Reid's 60th Birthday. [Online]. Available:
  \url{http://www.sciencedirect.com/science/article/pii/S0020740305000457}
\BIBentrySTDinterwordspacing

\bibitem{Mu2019}
X.~Mu, Y.~Xue, and Y.-b. Jia, ``{Robotic Cutting : Mechanics and Control of
  Knife Motion},'' pp. 3066--3072, 2019.

\bibitem{Long2014}
P.~Long, W.~Khalil, and P.~Martinet, ``{Force / vision control for robotic
  cutting of soft materials *}.''

\bibitem{Amos:2018:DME:3327757.3327922}
\BIBentryALTinterwordspacing
B.~Amos, I.~D.~J. Rodriguez, J.~Sacks, B.~Boots, and J.~Z. Kolter,
  ``Differentiable mpc for end-to-end planning and control,'' in
  \emph{Proceedings of the 32Nd International Conference on Neural Information
  Processing Systems}, ser. NIPS'18.\hskip 1em plus 0.5em minus 0.4em\relax
  USA: Curran Associates Inc., 2018, pp. 8299--8310. [Online]. Available:
  \url{http://dl.acm.org/citation.cfm?id=3327757.3327922}
\BIBentrySTDinterwordspacing

\bibitem{mitsioni2019data}
I.~Mitsioni, Y.~Karayiannidis, J.~A. Stork, and D.~Kragic, ``Data-driven model
  predictive control for the contact-rich task of food cutting,'' \emph{2019
  IEEE-RAS International Conference on Humanoid Robots (Humanoids)}, 2019.

\bibitem{deepmpc}
I.~Lenz, R.~A. Knepper, and A.~Saxena, ``Deepmpc: Learning deep latent features
  for model predictive control.'' in \emph{Robotics: Science and Systems},
  2015.

\bibitem{Jung1999}
S.~Jung and T.~Hsia, ``Adaptive force tracking impedance control of robot for
  cutting nonhomogeneous workpiece,'' vol.~3, 02 1999, pp. 1800 -- 1805 vol.3.

\bibitem{Zeng2002}
G.~Zeng and A.~Hemami, ``{An adaptive control strategy for robotic cutting},''
  no. April, pp. 22--27, 2002.

\bibitem{Sharma2019}
\BIBentryALTinterwordspacing
M.~Sharma, K.~Zhang, and O.~Kroemer, ``{Learning Semantic Embedding Spaces for
  Slicing Vegetables},'' 2019. [Online]. Available:
  \url{http://arxiv.org/abs/1904.00303}
\BIBentrySTDinterwordspacing

\bibitem{billard1}
K.~Kronander and A.~Billard, ``Learning compliant manipulation through
  kinesthetic and tactile human-robot interaction,'' \emph{Haptics, IEEE
  Transactions on}, vol.~7, pp. 367--380, 07 2014.

\bibitem{Huang2016}
\BIBentryALTinterwordspacing
B.~Huang, M.~Li, R.~L. De~Souza, J.~J. Bryson, and A.~Billard, ``A modular
  approach to learning manipulation strategies from human demonstration,''
  \emph{Autonomous Robots}, vol.~40, no.~5, pp. 903--927, Jun 2016. [Online].
  Available: \url{https://doi.org/10.1007/s10514-015-9501-9}
\BIBentrySTDinterwordspacing

\bibitem{deformable}
A.~X. {Lee}, H.~{Lu}, A.~{Gupta}, S.~{Levine}, and P.~{Abbeel}, ``Learning
  force-based manipulation of deformable objects from multiple
  demonstrations,'' in \emph{2015 IEEE International Conference on Robotics and
  Automation (ICRA)}, May 2015, pp. 177--184.

\bibitem{peginhole}
T.~Tang, H.-C. Lin, and M.~Tomizuka, ``A learning-based framework for robot
  peg-hole-insertion,'' in \emph{ASME 2015 Dynamic Systems and Control
  Conference}, 10 2015, p. V002T27A002.

\bibitem{Yfan}
Y.~{Fan}, J.~{Luo}, and M.~{Tomizuka}, ``A learning framework for high
  precision industrial assembly,'' in \emph{2019 International Conference on
  Robotics and Automation (ICRA)}, May 2019, pp. 811--817.

\bibitem{Johannink2018ResidualRL}
T.~Johannink, S.~Bahl, A.~Nair, J.~Luo, A.~Kumar, M.~Loskyll, J.~A. Ojea,
  E.~Solowjow, and S.~Levine, ``Residual reinforcement learning for robot
  control,'' \emph{2019 International Conference on Robotics and Automation
  (ICRA)}, pp. 6023--6029, 2018.

\bibitem{Bengio:2009:CL:1553374.1553380}
\BIBentryALTinterwordspacing
Y.~Bengio, J.~Louradour, R.~Collobert, and J.~Weston, ``Curriculum learning,''
  in \emph{Proceedings of the 26th Annual International Conference on Machine
  Learning}, ser. ICML '09.\hskip 1em plus 0.5em minus 0.4em\relax New York,
  NY, USA: ACM, 2009, pp. 41--48. [Online]. Available:
  \url{http://doi.acm.org/10.1145/1553374.1553380}
\BIBentrySTDinterwordspacing

\bibitem{Ebert2018}
\BIBentryALTinterwordspacing
F.~Ebert, S.~Dasari, A.~X. Lee, S.~Levine, and C.~Finn, ``{Robustness via
  Retrying: Closed-Loop Robotic Manipulation with Self-Supervised Learning},''
  no. CoRL, pp. 1--12, 2018. [Online]. Available:
  \url{http://arxiv.org/abs/1810.03043}
\BIBentrySTDinterwordspacing

\bibitem{Xu2017}
\BIBentryALTinterwordspacing
D.~Xu, S.~Nair, Y.~Zhu, J.~Gao, A.~Garg, L.~Fei-Fei, and S.~Savarese, ``{Neural
  Task Programming: Learning to Generalize Across Hierarchical Tasks},'' 2017.
  [Online]. Available: \url{http://arxiv.org/abs/1710.01813}
\BIBentrySTDinterwordspacing

\bibitem{DBLP:journals/corr/RanzatoCAZ15}
\BIBentryALTinterwordspacing
M.~Ranzato, S.~Chopra, M.~Auli, and W.~Zaremba, ``Sequence level training with
  recurrent neural networks,'' in \emph{4th International Conference on
  Learning Representations, {ICLR} 2016, San Juan, Puerto Rico, May 2-4, 2016,
  Conference Track Proceedings}, 2016. [Online]. Available:
  \url{http://arxiv.org/abs/1511.06732}
\BIBentrySTDinterwordspacing

\bibitem{maxplank3}
S.~Sahoo, C.~Lampert, and G.~Martius, ``{Learning Equations for Extrapolation
  and Control},'' \emph{ArXiv e-prints}, jun 2018.

\bibitem{Jonschkowski2017}
\BIBentryALTinterwordspacing
R.~Jonschkowski, R.~Hafner, J.~Scholz, and M.~Riedmiller, ``{PVEs:
  Position-Velocity Encoders for Unsupervised Learning of Structured State
  Representations},'' 2017. [Online]. Available:
  \url{http://arxiv.org/abs/1705.09805}
\BIBentrySTDinterwordspacing

\bibitem{Siciliano:2007:SHR:1209344}
B.~Siciliano and O.~Khatib, \emph{Springer Handbook of Robotics}.\hskip 1em
  plus 0.5em minus 0.4em\relax Berlin, Heidelberg: Springer-Verlag, 2007.

\bibitem{maaten2008visualizing}
L.~v.~d. Maaten and G.~Hinton, ``Visualizing data using t-sne,'' \emph{Journal
  of machine learning research}, vol.~9, no. Nov, pp. 2579--2605, 2008.

\bibitem{Cotin2000}
\BIBentryALTinterwordspacing
S.~Cotin, H.~Delingette, and N.~Ayache, ``{A hybrid elastic model for real-time
  cutting, deformations, and force feedback for surgery training and
  simulation},'' \emph{The Visual Computer}, vol.~16, no.~8, pp. 437--452,
  2000. [Online]. Available: \url{http://link.springer.com/10.1007/PL00007215}
\BIBentrySTDinterwordspacing

\bibitem{Chung2014EmpiricalEO}
J.~Chung, Çaglar G{\"u}lçehre, K.~Cho, and Y.~Bengio, ``Empirical evaluation
  of gated recurrent neural networks on sequence modeling,'' \emph{ArXiv}, vol.
  abs/1412.3555, 2014.

\bibitem{RHC}
D.~Q. {Mayne} and H.~{Michalska}, ``Receding horizon control of nonlinear
  systems,'' \emph{IEEE Transactions on Automatic Control}, vol.~35, no.~7, pp.
  814--824, July 1990.

\bibitem{adam}
D.~Kingma and J.~Ba, ``Adam: A method for stochastic optimization,''
  \emph{International Conference on Learning Representations}, 12 2014.

\bibitem{shooting}
A.~Rao, ``A survey of numerical methods for optimal control,'' \emph{Advances
  in the Astronautical Sciences}, vol. 135, 01 2010.

\end{thebibliography}

\end{document}